\documentclass[10pt,twocolumn,letterpaper]{article}

\usepackage{cvpr}              

\definecolor{cvprblue}{rgb}{0.21,0.49,0.74}
\usepackage[pagebackref,breaklinks,colorlinks,allcolors=cvprblue]{hyperref}
\usepackage{tcolorbox}
\usepackage{multirow}
\tcbset{colback=white}
\colorlet{titleblue}{blue!80!black}
\colorlet{titlered}{red!80!black}
\colorlet{titlegreen}{green!80!black}

\def\paperID{5984} 
\def\confName{CVPR}
\def\confYear{2025}

\title{On-Board Vision-Language Models for Personalized Autonomous Vehicle Motion Control: System Design and Real-World Validation}

\author{%
  Can Cui$^1$ , Zichong Yang$^1$ , Yupeng Zhou$^1$ , Juntong Peng$^1$, \\Sung-Yeon Park$^{1}$, Cong Zhang$^{1}$, Yunsheng Ma$^1$, Xu Cao$^2$, Wenqian Ye$^3$, \\Yiheng Feng$^1$, Jitesh Panchal$^{1}$, Lingxi Li$^{1}$, Yaobin Chen$^{1}$, Ziran Wang$^{1}$\\
  $^{1}$Purdue University\quad
  $^{2}$University of Illinois Urbana-Champaign\quad
  $^{3}$University of Virginia \\
  \texttt{\small \{cancui, ziran\}@purdue.edu}
}

\begin{document}
\maketitle
\begin{abstract}

Personalized driving refers to an autonomous vehicle's ability to adapt its driving behavior or control strategies to match individual users' preferences and driving styles while maintaining safety and comfort standards. However, existing works either fail to capture every individual preference precisely or become computationally inefficient as the user base expands. Vision-Language Models (VLMs) offer promising solutions to this front through their natural language understanding and scene reasoning capabilities. In this work, we propose a lightweight yet effective on-board VLM framework that provides low-latency personalized driving performance while maintaining strong reasoning capabilities. Our solution incorporates a Retrieval-Augmented Generation (RAG)-based memory module that enables continuous learning of individual driving preferences through human feedback. Through comprehensive real-world vehicle deployment and experiments, our system has demonstrated the ability to provide safe, comfortable, and personalized driving experiences across various scenarios and significantly reduce takeover rates by up to 76.9\%. To the best of our knowledge, this work represents the first end-to-end VLM-based motion control system in real-world autonomous vehicles.


\end{abstract}    
\section{Introduction}
\label{sec:intro}

The autonomous driving industry is experiencing an evolution towards human-centric systems~\cite{calvert2020human,lou2024human}, where vehicle automation extends beyond only considering traditional safety and efficiency metrics but also considers understanding users' implicit instructions and providing personalized driving experiences~\cite{cui2024personalizedautonomousdrivinglarge,cui2024largelanguagemodelsautonomous}. Personalized driving experiences are crucial for user acceptance and trust, as they help bridge the gap between autonomous technology and human expectations. This trend reflects a growing recognition that successful adoption of the autonomous vehicle requires not just technically self-driving, but also the ability to provide human-like driving experiences that align with individual preferences and expectations~\cite{cui2024personalizedautonomousdrivinglarge,cui2024largelanguagemodelsautonomous}. 

Previous work in personalized autonomous driving has followed two main approaches. The first uses clustering algorithms to classify drivers into broad categories (e.g., aggressive or conservative), but this fails to capture individual nuances and preferences, forcing users into predefined groups that may not match their actual driving style~\cite{wang2021personalized,wang2020driver}. The second approach develops individual models for each user through learning-based methods~\cite{du2023driver,wang2022gaussian}, but this requires extensive training data per user and becomes computationally inefficient as the user base grows. Furthermore, these methods lack the ability to reason about real-time human instructions or adapt to changing environments.

Recent advances in Vision-Language Models (VLMs) have demonstrated promising capabilities in understanding complex driving scenarios and natural language instructions through their integration of computer vision and language processing~\cite{ma2023dolphinsmultimodallanguagemodel, Park_2024_WACV,Cui_2024_WACV,chen2024driving}.  The development in VLMs has led researchers to leverage VLMs' multimodal understanding capabilities to enhance both perception and decision-making in autonomous systems~\cite{tian_drivevlm_2024,pan_vlp_2024}. However, there remains a research gap in leveraging VLMs to enhance control policies or adapt them to individual driving preferences and styles. This gap is particularly evident in the challenge of translating a high-level understanding of human preferences and scenario information into actionable low-level control policies. Additionally, the computational demands of previously adopted large-scale models make on-board deployment infeasible, forcing reliance on cloud-based inferencing. This solution depends on stable internet connectivity and can introduce significant latency issues in the inference process, with response times reaching up to 3 or 4 seconds~\cite{pmlr-v202-liu23am}. This is incompatible with the reliable and near real-time requirements of autonomous driving.

To address these limitations, we propose a novel VLM-based framework for real-time personalized autonomous vehicle control. Our system enables efficient on-board deployment while maintaining strong instruction understanding and scene reasoning capabilities. We present the \textbf{first-of-its-kind} real-world implementation of an on-board VLM-based personalized motion control system. The main contributions of this work are:

\begin{itemize}
\item We develop an efficient on-board VLM that achieves comparable reasoning capabilities to cloud-based solutions while operating independently of internet connectivity. Our lightweight solution addresses critical computational constraints for real-world autonomous vehicles.
\item We propose a novel approach to translate diverse human instructions and visual inputs into actionable control policies, handling both explicit commands (`go faster') and implicit feedback (`I feel uncomfortable') and diverse environment conditions.
\item We introduce a RAG-based memory module that incorporates human feedback for continuous learning and adaptation, enabling personalized driving experiences through iterative refinement of control strategies.
\item Through extensive real-world deployment and experiments, we demonstrate safe, comfortable, and reliable autonomous driving performance across various instruction types and successful personalization capabilities, reducing takeover rates by up to 76.9\%.
\end{itemize}

\section{Related Works}
\label{sec:related}

\subsection{Vision-Language Models for Autonomous Driving}
Early applications of language models in autonomous driving involved human-guided planning, integrating natural language commands or advice, and generating language-based interpretations or control signals for vehicle operations \cite{kim2019groundinghumantovehicleadviceselfdriving, kim2020advisable,long2024vlm}. With the advent of VLMs, initial efforts focused on image-based models \cite{li2023blip2bootstrappinglanguageimagepretraining, liu2023visualinstructiontuning, bai2023qwenvlversatilevisionlanguagemodel, dai2023instructblipgeneralpurposevisionlanguagemodels} that utilized image encoders and bridging modules connected to LLMs. More recently, video-based VLMs have emerged \cite{luo2023valleyvideoassistantlarge, zhang2023videollamainstructiontunedaudiovisuallanguage, li2024videochatchatcentricvideounderstanding}, enhancing their applicability in autonomous driving by processing image sequences crucial for real-time decision-making. Fine-tuned with instruction-following datasets specific to driving scenes, these VLMs address tasks such as visual question answering (VQA) to achieve a comprehensive understanding of traffic scenarios and the behavior of the ego-vehicle \cite{ma2023dolphinsmultimodallanguagemodel, Park_2024_WACV}. Additionally, predicting future waypoints has become a prominent task within this domain, often employing Chain-of-Thought mechanisms to improve planning by generating text sequences for perception, prediction, and planning \cite{sima2024drivelmdrivinggraphvisual, tian2024drivevlmconvergenceautonomousdriving}. Certain models do not solely rely on image inputs; instead, they incorporate perception and prediction submodules to enrich prompts with detailed road agent information for more effective planning \cite{pan2024vlpvisionlanguageplanning}. Moreover, some VQA datasets include localized object data to better anticipate the future behavior of potential risk objects \cite{nie2024reason2driveinterpretablechainbasedreasoning, qian2024nuscenesqamultimodalvisualquestion}. Unlike existing works that primarily focus on VLMs operating in simulation environments, our approach addresses generating action policies that can be adapted to real-world vehicle-level applications using VLMs.

\subsection{Personalization/Human-AI Alignment in Autonomous Driving}

There has been an increased focus on the personalization of autonomous driving, aiming to follow the driving experience to individual preferences and needs~\cite{bae2022selfdriving,9419761}. In recent developments, various studies have explored personalized adaptive cruise control systems, focusing on steady-state operation~\cite{zhao2023personalized}, Gaussian process-based methods~\cite{wang2022gaussian,wang2021personalized}, Transformer and RNN integration~\cite{sachdeva2022gapformer} for enhanced user experience and driving preferences. Schrum et al. introduced MAVERIC, a novel framework for personalizing driving styles, which demonstrated the ability to adapt to humans' driving preferences influenced by personality traits and perceived similarity~\cite{schrum2023maveric}.  Buzdugan et al. focused on safety-critical scenarios, using driver-in-the-loop simulations to detect and adapt to personalized driving styles, thus bridging the gap between human driving behavior and autonomous vehicle predictability~\cite{vehicles5030064}. Ma et al.~\cite{ma2024lampilot} leveraged RAG which is an approach that enhances model capabilities by retrieving relevant historical information to augment the generation process\cite{lewis2020retrieval} to learn from human feedback. There are also studies on systems offering personalized experiences to predict human actions and increase human trust instead of directly altering vehicle control~\cite{zhang2023evaluation,du2023driver,wang2022mobility,liao2023driver,sun2020exploring}. However, such personalization frameworks often encounter limitations such as dynamically adapting to human preferences or unseen traffic scenarios. This is where VLMs could potentially complement these systems by offering more complex and context-aware adaptations, leveraging their advanced language understanding and generative capabilities.

\section{On-Board Vision-Language Models for Personalized Motion Control}
\begin{figure*}[!t]
    \centering
    \includegraphics[width=0.9\textwidth]{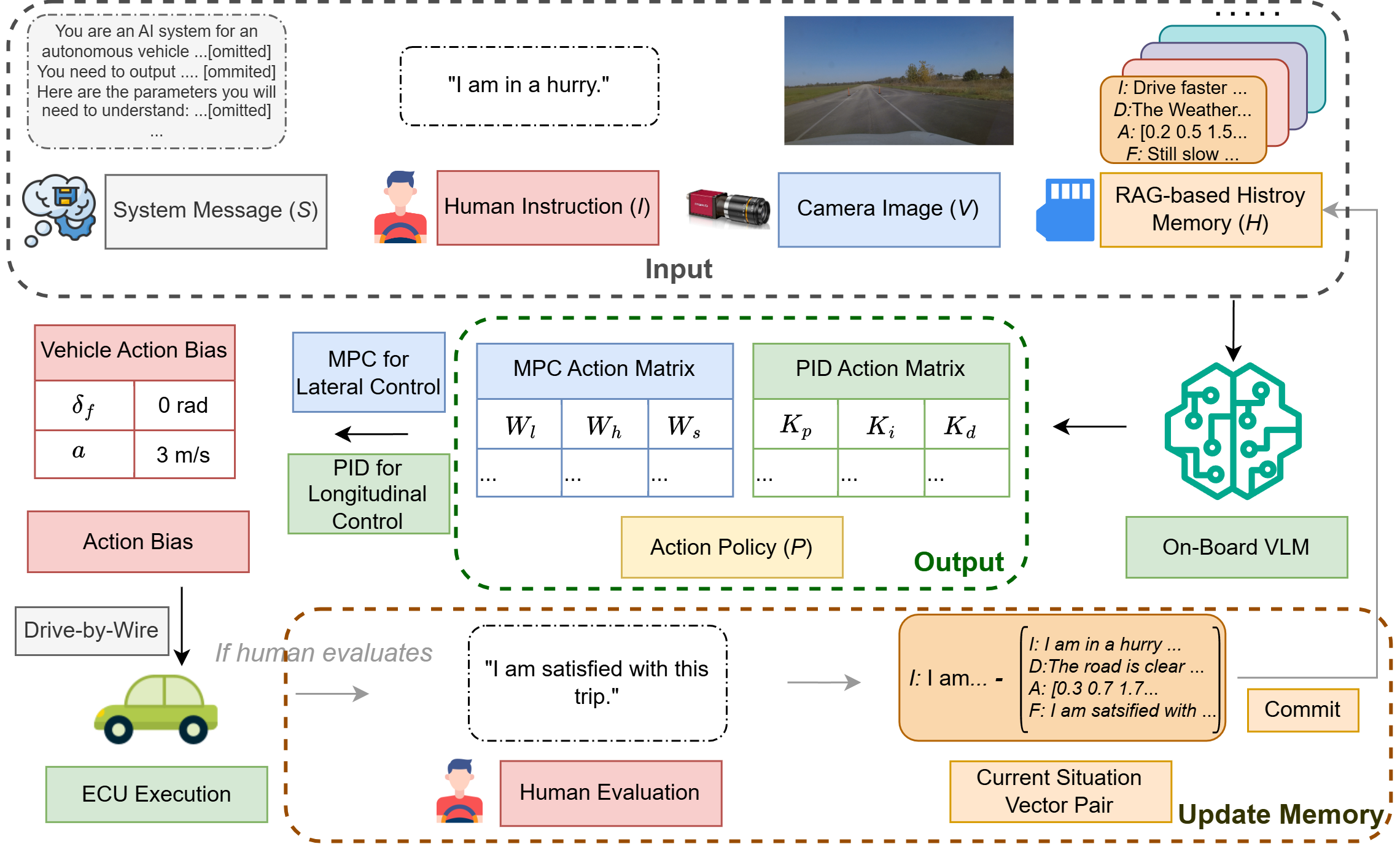}
    \caption{An overview of the proposed framework for personalized autonomous vehicle motion control. The system workflow begins with processing four input streams (System Message $S$, Human Instruction $I$, Camera Image $V$, and Historical Memory $H$) through an on-board VLM, which generates personalized action policies $P$ containing MPC and PID control parameter matrices. These policies are then executed through the vehicle's drive-by-wire system. If the human evaluates, human feedback $F$ is collected and stored in the RAG-based memory module for continuous learning and adaptation of the system's behavior to individual preferences.}
    \vspace{-5mm}
    \label{fig:main}
\end{figure*}
\label{sec:method}


In this section, we present our on-board VLM for personalized motion control in autonomous driving, designed to accommodate individual driving styles. Our approach leverages a compact 9B-parameter VLM, fine-tuned from Qwen-VL~\cite{bai2023qwenvlversatilevisionlanguagemodel}, which processes both visual information (including weather conditions, road types, and traffic situations) and verbal commands to generate personalized control strategies for each user. The reduced scale of this VLM enables edge deployment while maintaining command interpretation and reasoning capabilities, allowing the system to effectively understand and respond to implicit human instructions. The overall framework is shown in Fig. \ref{fig:main}.

\subsection{Problem Statement}
  
In this paper, we adopt the traditional module-based autonomous driving framework that includes the full pipeline from perception to motion control, and our focus is specifically on enhancing the decision-making process at the motion control level, adapting the control performance to accommodate different human driving styles. The goal of the proposed system is to translate both verbal commands $\boldsymbol{I}$ and visual inputs $\boldsymbol{V}$ into executable control sequences for the motion control process. The onboard VLM acts as a translation mechanism $f:(\boldsymbol{I},\boldsymbol{V})\rightarrow\boldsymbol{P}$ that generates a policy $\boldsymbol{P}$, which is then fed into predefined maneuver programs.

Additionally, system messages $\boldsymbol{S}$ are sent to our VLM to specify both the tasks and adjustment strategies. In practice, $\boldsymbol{S}$ is generated through a predefined language generator. These system messages define the VLM's role and objectives, and provide explicit instructions for the tuning strategy.


Simultaneously, to further enhance personalization, we implement a RAG system called the memory module to build a database storing historical human-vehicle interactions. Whenever a human activates this system, only relevant historical scenarios $\boldsymbol{H}$ are retrieved and provided to the VLM as reference. After each trip, users can provide feedback $\boldsymbol{F}$ on the generated control policy $\boldsymbol{P}$ for the current situations (including instructions $\boldsymbol{I}$ and visual input $\boldsymbol{V}$), which helps refine the VLM's reasoning process. Subsequently, the instructions $\boldsymbol{I}$, scene description $\boldsymbol{D}$, policy $\boldsymbol{P}$, and feedback  $\boldsymbol{F}$ are packaged as a historical data entry and stored in the RAG database. Therefore, there are three procedures in our system:
\vspace{-2mm}
\begin{equation}
    \begin{aligned}
        \text{VLM Execution}:\quad & P \xleftarrow{VLM} f(I,S,V,H); \\
        \text{Human Feedback}:\quad & F \xleftarrow{\text{Human}} [I,V,P] ; \\
        \text{Memory Update}:\quad & H \leftarrow \left[I,D,P,F\right]
    \end{aligned}
\end{equation}

\subsection{System Input}

As illustrated in Fig. \ref{fig:main}, our fine-tuned on-board VLM processes four critical inputs for reasoning. The primary inputs consist of visual data $\boldsymbol{V}$ from the onboard camera and natural language commands $\boldsymbol{I}$ which are converted from verbal human instructions $\boldsymbol{I}$ using the open-source local speech recognition model `Whisper~\cite{radford2022robustspeechrecognitionlargescale}.' Notably, due to the advanced reasoning and understanding ability of our fine-tuned VLM, our system can interpret implicit expressions from humans such as ``It is nice weather, I want to enjoy the view." This ability to understand implicit instructions is crucial, as users typically communicate through natural conversational phrases rather than explicit value-based commands containing exact parameters. Furthermore, our system leverages contextual and environmental information captured in the visual inputs $\boldsymbol{V}$, including weather conditions, traffic situations, and road characteristics. For instance, the system automatically adopts a more conservative driving policy during nighttime operations or adverse weather conditions. 

Additionally, a pre-defined system message generator is employed to produce a customized system message $\boldsymbol{S}$, which is then simultaneously sent to the VLM. This message includes essential information about the system, including the user's identity, specific objectives, and key principles guiding the system’s behavior, particularly how to utilize the controller or tune parameters. 

Furthermore, the VLM incorporates relevant interaction history $\boldsymbol{H}$ extracted from our memory module as contextual input, which includes previous human instructions $\boldsymbol{I}$, scene descriptions $\boldsymbol{D}$, executed actions $\boldsymbol{A}$, and user feedback $\boldsymbol{F}$. This historical context enables the VLM to generate more appropriate responses by considering past interactions and human feedback. For example, if a user has previously expressed a preference for cautious driving in certain scenarios, the system can capture this preference into its current decision-making process, ensuring more personalized and contextually appropriate responses. A detailed discussion of how our memory module works will be presented in subsection \ref{subsec:mm}.


\subsection{Reasoning and Action Generation}



In our approach, reasoning within the VLM framework enables the interpretation of diverse driving scenarios and user instructions to generate actionable outputs. Traditional controllers in motion control typically rely on a default set of parameters; however, following the approach in \cite{sha2023languagempclargelanguagemodels}, our VLM will generate two distinct action matrices to separately manage the PID controller for longitudinal movement and the MPC for lateral movement. These matrices translate the model’s understanding of the environment and user preferences into precise control actions, guiding the autonomous vehicle's behavior. Specifically, they are used by the controllers to generate acceleration $a$ and steering angle $\delta_f$, which are executed by the vehicle’s ECU. The ECU then sends low-level control signals to the drive-by-wire system developed by AutonomousStuff~\cite{autostuff}, enabling smooth and responsive vehicle operation. The general process of this subsection can be shown below:
\begin{equation}
    \begin{aligned}
        \text{Output Action } &P:= 
        \begin{bmatrix}
            K_p & K_i &K_d \\
            W_l & W_h & W_s
        \end{bmatrix}    \\
        \text{Action Execution }& P  \xrightarrow{Controllers}\left[\delta_f,a\right] \rightarrow ECU
    \end{aligned}
\end{equation}


\subsection{RAG-Enhanced Memory Module}
\label{subsec:mm}



Given that our 8B-parameter VLM lacks the extensive reasoning capabilities of larger, 100B-200B parameter models, we employ a RAG-based approach~\cite{lewis2020retrieval} and integrate a memory module to enhance reasoning and enable human feedback learning. This system is built upon the Chroma vector database~\cite{chroma_ai-native_2023}, enabling efficient storage history interactions and retrieval of similar driving scenarios.

The memory module is uniquely created for each user, ensuring a personalized driving experience follows individual preferences and patterns. It stores driving-related information in a structured format comprising commands paired with corresponding context tuples: 
\begin{equation}
    \{I - ( I, D, P, F)\}
\end{equation}
When processing a new driving scenario, the instruction $\boldsymbol{I}$ is used for similarity matching to retrieve the top-k similar prior situations. The associated data values are then sent to the VLM, enhancing decision-making with relevant context and supporting personalization. This RAG-enhanced memory enables the VLM to handle similar situations with greater accuracy and consistency, improving the vehicle’s responsiveness to unique user preferences and enhancing the overall driving experience.

\subsection{Multi-Controller Joint Motion Control}

As shown in Fig. \ref{fig:main}, we implement a decoupled control strategy that separates lateral and longitudinal vehicle motion control. The lateral control is handled by MPC calculating the longitudinal acceleration $\alpha$, while longitudinal control is managed through a PID controller calculating the front steering angle $\delta_f$. The motion planning module in the upper layer provides trajectories consisting of waypoints, which our motion control system tracks. The calculated $\alpha$ and $\delta_f$ are then transmitted to the drive-by-wire system developed by AutonomousStuff~\cite{autostuff} for precise control of throttle, braking, and steering.

For the longitudinal control, the PID controller calculates the $\alpha$ for each time step $\Delta t$ to minimize the velocity error $e_v$, which is the difference between the current velocity $V_{\text{current}}$ and the desired velocity $V_{\text{ref}}$. 
 \begin{equation}
     \begin{aligned}
      \alpha(t) = K_p e_v(t) + K_i \sum_{i=0}^{t} e_v(i) \Delta t + K_d \frac{\Delta e_v(t)}{\Delta t}
      \end{aligned}
 \end{equation}
where $K_p$, $K_i$, and $K_d$ are the proportional terms, integration terms, and derivative terms that will be contained in the action matrix generated by our VLM.

For the lateral control, our MPC approach utilizes a linear vehicle dynamics model \cite{snider2009automatic} to predict future states and optimize the front steering angle, \(\delta_f\), over a finite prediction horizon. With the prediction model\cite{snider2009automatic}, the control increment is then obtained by solving a Quadratic Program (QP) \cite{bemporad2000explicit} to minimize the cost function \(J\) in the MPC:
\begin{equation}
    J = E^T Q E + \Delta_f^T R \Delta_f
\end{equation}
where \(E\) is the predicted state calculated by the prediction model, and \(\Delta_f\) is the future control input. The \(Q\) and \(R\) are weighting matrices that penalize tracking state deviations and control effort. Our VLM generates the action matrix that primarily considers three key components: the weight for lateral error ($W_l \in Q$), the weight for heading error ($W_h \in Q$), and the weight for the squared terms of speed and steering inputs ($W_s \in R$). These weighting factors are selected as they demonstrate the most significant impact on lateral control performance.

\subsection{Efficient On-Board VLM Module} 

We generate a dataset of 10,000 image-instruction pairs, each labeled with the desired action, to create a comprehensive training set for fine-tuning our on-board VLM. This VLM is based on the Qwen-VL architecture~\cite{bai2023qwenvlversatilevisionlanguagemodel}, which we fine-tune using the Low-Rank Adaptation (LoRA) method~\cite{hu2021loralowrankadaptationlarge} (a type of Parameter Efficient Fine-Tuning (PEFT)~\cite{xu2023parameterefficientfinetuningmethodspretrained}), enabling significant customization while preserving computational efficiency. To optimize for on-board deployment, we apply 4-bit Activation-Aware Weight Quantization (AWQ)~\cite{lin2024awqactivationawareweightquantization}, compressing the VLM to increase inference speed without sacrificing too much accuracy. This combination of techniques ensures a responsive, on-board VLM suited to real-time response.

\paragraph{Dataset Collection}

We develop a specialized training dataset to fine-tune the Qwen-VL model~\cite{bai2023qwenvlversatilevisionlanguagemodel}, consisting of 1,200 semi-human-annotated image-text pairs. Each image, representing a traffic scene sourced from the NuScenes dataset, which includes numerous diverse traffic scenarios, is paired with a human-provided instruction and a corresponding action label in the form of a controller action matrix, guiding the model’s response in different traffic scenarios.

The human instructions are also very diverse, ranging from explicit commands like `speed up' to more implicit cues such as `I am in an urgent situation.' This diversity allows the VLM to interpret both clear and vague inputs, improving its ability to understand complex human intentions. To enhance the model's responsiveness to different driving styles, we annotate each image with three different instruction types—aggressive, moderate, and conservative—each paired with a corresponding action. This multi-faceted approach ensures that the VLM can adapt its behavior to match various driving styles, enabling it to respond flexibly and contextually across diverse traffic conditions.


\paragraph{LoRA Finetune}

We apply the LoRA method to fine-tune our Qwen-VL model. LoRA works by freezing the pre-trained model weights and introducing trainable, low-rank decomposition matrices into each layer of the Transformer architecture. This approach significantly reduces the number of trainable parameters required, making fine-tuning more efficient.

The fine-tuning process for our VLM is conducted on a cluster of four NVIDIA A100 GPUs, each equipped with 40GB of memory. The model is trained over five epochs with a per-device batch size of two for training and one for validation, using a learning rate of 1e-5. Additionally, we implemented gradient accumulation with eight steps, allowing for effective larger batch processing. This setup enables the entire training process to be completed in approximately five hours, ensuring both accuracy and efficiency in model performance.

\paragraph{Compression and On-Board Deployment of VLM}

AWQ~\cite{lin2024awqactivationawareweightquantization} is a hardware-friendly technique for low-bit, weight-only quantization, specifically designed for VLM. AWQ minimizes quantization error by identifying the 1\% salient weights, which are then scaled using an equivalent transformation to preserve their precision. We apply AWQ to quantize our model to INT4, achieving improved quantization performance suited for on-board deployment. Additionally, we utilize the LMDeploy toolkit~\cite{2023lmdeploy} to optimize inference time. This enhancement is made possible through features such as persistent batching, blocked KV cache, tensor parallelism, and optimized CUDA kernels, all of which contribute to high-performance, low-latency operation.

\section{Real-World Experiment}
\label{sec:exper}
To comprehensively evaluate our system's performance, we conduct a series of experiments assessing its ability to provide safe, comfortable, reliable, and personalized driving experiences. We employ multiple evaluation metrics: a driving score to measure driving performance, including safety, comfort, and alignment with environmental conditions and human instructions; takeover frequency to assess personalization capabilities; and evaluator-based assessments to investigate trustworthiness, reliability, and user satisfaction. Additionally, we perform an ablation study to examine the effectiveness of the memory module.

\begin{table*}[ht!]
\caption{Driving Performance Validation. $\downarrow$: Lower Values are Better. $\uparrow$: Higher Values are Better.}
\label{tab:drivingperformance}
\centering
\resizebox{2\columnwidth}{!}{%
\begin{tabular}{c|c|ccc|cccc|c|cc|c}
\toprule
\multirow{3}{*}{\begin{tabular}[c]{@{}c@{}}Driving \\ Scenario\end{tabular}} & \multirow{3}{*}{Model} & \multicolumn{3}{c|}{Safety Metrics} & \multicolumn{4}{c|}{Comfort Metrics} & Time Efficiency & \multicolumn{2}{c|}{Alignment} & \multirow{3}{*}{Driving Score$\uparrow$} \\ \cline{3-12}
&  & \begin{tabular}[c]{@{}c@{}}Time to \\ Collision($s$)$\uparrow$\end{tabular} & \begin{tabular}[c]{@{}c@{}}$SV_{x}$ \\ ($m^2/s^2$)$\downarrow$\end{tabular} & \begin{tabular}[c]{@{}c@{}}$SV_{y}$ \\ ($10^{-2} m^2/s^2$)$\downarrow$\end{tabular} & \begin{tabular}[c]{@{}c@{}}$|\bar{\alpha_{x}}|$\\ ($m/s^2$)$\downarrow$\end{tabular} & \begin{tabular}[c]{@{}c@{}}$|\bar{J_{x}}|$\\ ($m/s^3$)$\downarrow$\end{tabular} & \begin{tabular}[c]{@{}c@{}}$|\bar{\alpha_{y}}|$\\ ($10^{-1} m/s^2$)$\downarrow$\end{tabular} & \begin{tabular}[c]{@{}c@{}}$|\bar{J_{y}}|$\\ ($m/s^3$)$\downarrow$\end{tabular} & \begin{tabular}[c]{@{}c@{}}LLM \\ Latency($s$)$\downarrow$\end{tabular} & \begin{tabular}[c]{@{}c@{}}Command\\ Alignment$\uparrow$\end{tabular} & \begin{tabular}[c]{@{}c@{}}Scenario\\ Alignment$\uparrow$\end{tabular} &  \\
\midrule
\multirow{3}{*}{Acceleration} & Baseline & 2.44& 28.8 & 0.36 & 0.78 & 3.27 & 0.46 & 0.44 & - & 92.0 & 60.0 & 75.6\\
& GPT-4o & 2.52 & 30.0 & 0.39 & 0.83 & 3.40 & 0.52 & 0.52 & 5.82 & 92.9 & \textbf{71.3} & 76.4\\
& \textit{Ours} & 2.46 & 30.8 & 0.39 & 0.81 & 3.07 & 0.53 & 0.81 & 1.98 & \textbf{96.3} & 60.9 & \textbf{76.5} \\
\midrule
\multirow{3}{*}{\begin{tabular}[c]{@{}c@{}}Lane\\ Change\end{tabular}} & \textit{Baseline} & 2.44 & 3.91 & 1.65 & 0.37 & 3.14 & 0.88 & 1.01 & - & 88.5 & 60.0 & 74.5\\
& GPT-4o & 2.71 & 3.88 & 2.23 & 0.53 & 4.38 & 1.13 & 1.39 & 4.84 & 90.4 & \textbf{88.6} & \textbf{78.4}\\
& \textit{Ours} & 2.15 & 4.07 & 2.15 & 0.41 & 3.35 & 0.98 & 1.02 & 1.83 & \textbf{92.2} & 71.9 & 77.5\\
\midrule
\multirow{3}{*}{\begin{tabular}[c]{@{}c@{}}Left\\ Turn\end{tabular}} & Baseline & - & 1.12 & 7.52 & 0.22 & 1.81 & 1.29 & 1.36 & - & 88.0 & 60.0 & 70.4\\
& GPT-4o & - & 0.93 & 11.5 & 0.29 & 2.75 & 2.11 & 2.40 & 5.23 & \textbf{91.3} & \textbf{85.0} & 71.4\\
& \textit{Ours} & - & 0.94 & 6.74 & 0.19 & 1.67 & 1.33 & 1.32 & 1.64 & 90.2 & 67.8 & \textbf{74.4}\\
\bottomrule
\end{tabular}%
}
\vspace{-5mm}
\end{table*}

\subsection{Experiment Setup}\label{subsec:setup}
\paragraph{Autonomous Vehicle Setup:}

\begin{figure}[t]
    \centering
    \includegraphics[width=\linewidth]{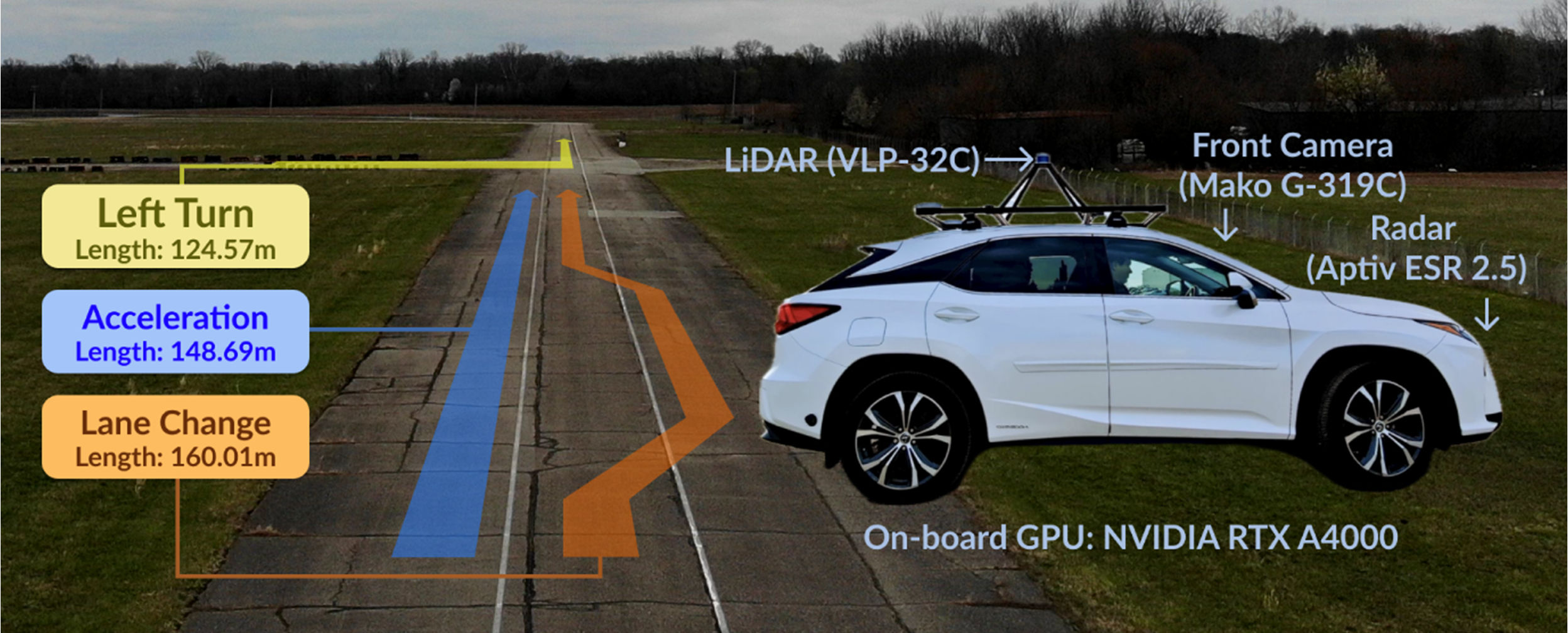}
    \caption{Overview of the experiment setup.}
    \vspace{-5mm}
    \label{fig:test_track}
\end{figure}

As shown in Fig. \ref{fig:test_track}, we use an autonomous vehicle to conduct real-world experiments, which is a drive-by-wire-enabled 2019 Lexus RX450h. We deploy the open-source autonomous driving software Autoware.AI~\cite{autoware} with ROS Melodic in Ubuntu 18.04. We use 3D-NDT~\cite{NDT_algorithm,3Dndt} for mapping and localization. An Aptiv ESR 2.5 radar, a Velodyne VLP-32C LiDAR, and two Mako-G319C cameras are deployed on the vehicle to enable the perception capabilities. The on-board computer has an Intel i9-9900 9th Gen 3.10/5.0GHz hexa-core 65W processor with eight cores and 16 threads, 64GB RAM, NVIDIA Quadro RTX-A4000 16GB GPU, and 512GB NVMe solid state drive.

\paragraph{Test Track and Participants}

The field experiments\footnote{The experiments conducted in this study satisfy all local traffic guidelines and guarantee the safety of the participants. A human always sits in the driver’s seat of the autonomous vehicle to monitor its status and get ready to take over.} aim at validating the real-world performance of our personalized motion control system. We include three types of driving behaviors—accelerating, lane changing, and turning—to comprehensively validate control over steering, throttle, and braking. An overview of the test track and driving behaviors is shown in Fig. \ref{fig:test_track}. For both acceleration and lane change scenarios, a lead vehicle is positioned 30 $m$ ahead of the ego vehicle, accelerating from static to 45 $km/h$ with an acceleration of 1.26 $m/s^2$. In the acceleration scenario, the ego vehicle accelerates from a complete stop to reach 50 $km/h$. In the lane change scenario, the ego vehicle maintains 50 $km/h$ while overtaking the lead vehicle. For the intersection turning scenario, the ego vehicle navigates a curve with a radius of 23.89 $m$ at a constant speed of 30 $km/h$.


Our study includes seven participants with diverse demographic characteristics. The participants consisted of 61.4\% male and 28.6\% female drivers, with ages ranging from 23 to 30 years (mean = 26.42, std = 3.24). Their driving experience varies considerably (mean = 6.42 years, std = 4.27 years). All participants hold valid U.S. driving licenses.

\paragraph{Instruction Directness}

In the field of linguistics, instructions can range from simple to complex in terms of how directly they convey intent~\cite{yule2022study}. To evaluate our model's natural language understanding capabilities, we classify instructions into three levels of increasing complexity: Level 1 consists of straightforward commands that explicitly state the desired action; Level 2 includes moderately complex instructions that may require some contextual understanding; Level 3 represents sophisticated commands that involve implicit meanings or complex conditions.





\subsection{System Driving Performance}

\begin{figure}[t]
    \centering
    \includegraphics[width=\linewidth]{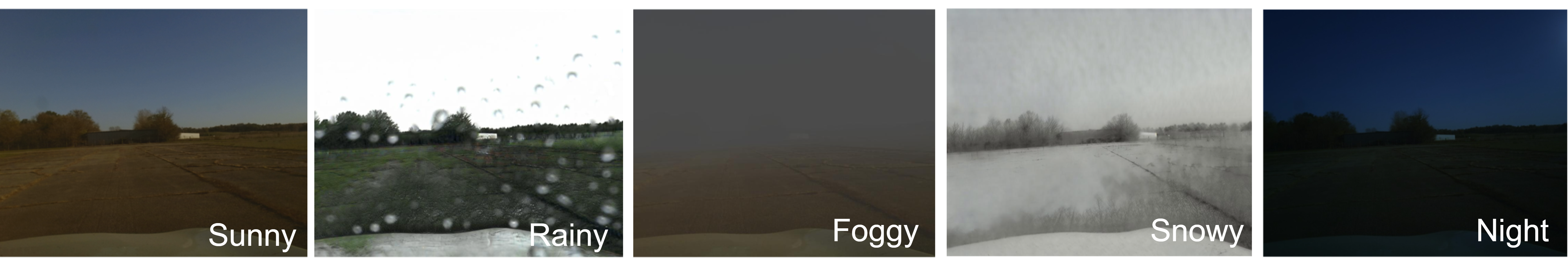}
    \caption{Sample vision inputs from different weather scenarios.}
    \vspace{-5mm}
    \label{fig:weather}
\end{figure}

To showcase our system's driving performance, we conduct comparative experiments against two systems: a baseline system using pre-defined controller parameters for general safety and comfort and a system utilizing GPT-4o with few-shot learning. We evaluate across three primary meta-driving scenarios (acceleration, lane changing, and turning), with each scenario tested under ten different commands and five weather conditions (sunny, rain, fog, snow, and night) to test the model's vision understanding as shown in Fig.~\ref{fig:weather}.

\paragraph{Evaluate Metrics} The models are then assessed based on four key aspects—safety, comfort, time efficiency, and alignment, as shown in Tab.~\ref{tab:drivingperformance}. An overall driving score $S$ is then calculated as a weighted sum of the individual metric scores, denoted as:
\begin{equation}
S = \sum{w_k \cdot S_k}
\end{equation}
where $k$ includes all ten metrics: Time to Collision ($\tau$), longitudinal and lateral speed variance ($SV_x$ and $SV_y$), lateral and longitudinal mean absolute acceleration ($|\bar{\alpha_{x}}|$ and $|\bar{\alpha_{y}}|$), lateral and longitudinal mean absolute jerk ($|\bar{J_{x}}|$ and $|\bar{J_{y}}|$), LLM Latency, Command Alignment and Scenario Alignment.  All the scores $S_k$ range from 0 to 100, while the weights of scores $w_k$ are empirically tuned for each driving scenario. For instance, longitudinal parameters have higher weights in acceleration scenarios, while lateral parameters are weighted more heavily in turning scenarios. For safety metrics, we set a critical Time to Collision threshold $\tau$ = 1.5 to prevent potential collisions. Other metrics like speed variance, acceleration, and jerk are scored relative to the baseline model. 

The alignment evaluation consists of two aspects. Command Alignment quantifies the deviation between the model output and the expected parameter range, calculated as a weighted average across six control parameters. For each parameter, we establish three ranges based on past experiments, corresponding to aggressive, conservative, or moderate driving styles.  Taking the PID controller's proportional parameter $K_p$ as an example, the score is calculated as:
\begin{equation}
\resizebox{0.4\textwidth}{!}{%
$
S_{K_{p}}^{*} = 
\begin{cases} 
\frac{100(K_p - K_{p, \min})}{K_{p, \text{lower}} - K_{p, \min}}, & K_p \in [K_{p, \min}, K_{p, \text{lower}}), \\[10pt]
100, & K_p \in [K_{p, \text{lower}}, K_{p, \text{upper}}), \\[10pt]
\frac{100(K_{p, \max} - K_p)}{K_{p, \max} - K_{p, \text{upper}}}, & K_p \in [K_{p, \text{upper}}, K_{p, \max}], \\[10pt]
0, & \text{else}.
\end{cases}%
$
}
\end{equation}
where $K_{p, \min}$ and $K_{p, \max}$ are the minimum and maximum overall parameter range obtained through experiments, while $K_{p, \text{lower}}$ and $K_{p, \text{upper}}$ are determined by the command intention labeled by human experts. The scenario alignment score computes whether the model can capture details of the scene through vision input and act more aggressively or conservatively based on the current condition. It is calculated by tallying the percentage of instances where the model gives more conservative parameter sets in adverse weather conditions compared to the sunny clear scenarios. 

\paragraph{Result} As shown in Tab.~\ref{tab:drivingperformance}, the command alignment score of our VLM model is similar to or greater than GPT-4o, showing our model has high reasoning capabilities regarding the personalization of control parameters. As for the scenario alignment, our model shows significantly better performance than baseline conditions in lane changing and left turn scenarios but a score very similar to the baseline scenario. We think this is mostly due to the model considering acceleration on a clear straight road in less dangerous situations and thus does not act too conservatively. In terms of the overall Driving Score, our model also surpasses the performance of baseline models and even GPT-4o in some scenarios, indicating our model provides the most well-rounded driving experience.

\subsection{Human-in-the-Loop Validation}

\begin{figure}[!t]
    \centering
    \includegraphics[width=0.9\linewidth]{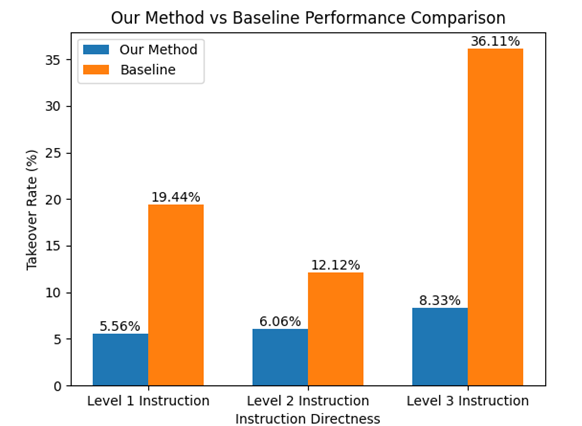}
    \caption{Takeover rate comparison between the baseline and our method.}
    \vspace{-7mm}
    \label{fig:takeover}
\end{figure}

\begin{figure*}[!t]
    \centering
    \includegraphics[width=\textwidth]{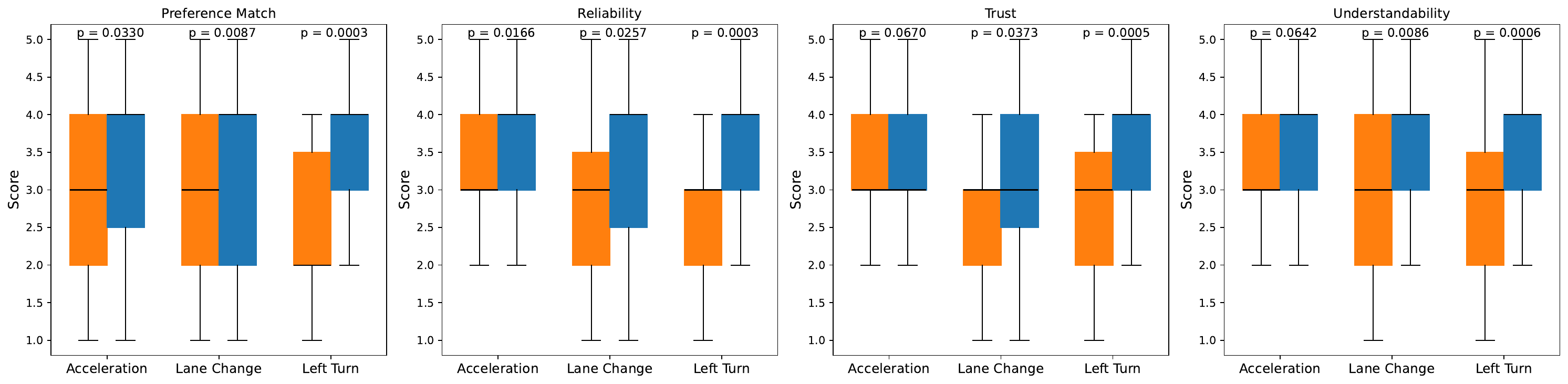}
    \caption{Median comparison between VLM-based model and the baseline under various driving scenarios. \textcolor{Orange}{Orange} represents baseline while \textcolor{Blue}{Blue} represents our method.}
    \vspace{-7mm}
    \label{fig:median}
\end{figure*}


This subsection evaluates the effectiveness of our method in reducing takeover rates compared to the baseline system. The baseline consists of the traditional autonomous vehicle controller with default settings, where two unified controllers manage vehicle operations on longitudinal and lateral respectively. We compare this conventional approach against our VLM-based adaptive motion control system. Throughout the experiments, participants could provide explicit instructions or implicit preferences/feedback with varying degrees of directness, prompting the system to make corresponding adjustments. The instructions were categorized into three levels of directness, as defined in Subsection \ref{subsec:setup}.

Every participant is supposed to provide at least five instructions for each scenario. For each instruction-scenario pair, participants completed two trips - one with the baseline system and one with our VLM solution. To ensure unbiased evaluation, participants are not informed which system they are using during each trip. We use the takeover rate as our primary metric to evaluate the system's ability to provide personalized driving experiences.


The results demonstrate that our VLM-based method consistently outperforms the baseline system across all three levels of instruction directness. With Level 1 (direct) instructions, our method achieves a 5.56\% takeover rate versus the baseline's 19.44\%, representing a 71.4\% reduction. For Level 2 (moderately direct) instructions, the rates are 6.06\% versus 12.12\%, showing a 50\% improvement. Most notably, for Level 3 (indirect) instructions, our method achieves 8.33\% versus the baseline's 36.11\%, marking a 76.9\% reduction in takeover rate. The result demonstrates our system's better capability in interpreting and acting based on user preferences, regardless of how explicitly they are communicated.


\subsection{Evaluator-based Assessment}

To assess the impact of our system on trust, reliability, personalization, and understandability, we conduct a survey to capture participants' attitudes from various perspectives, rated on a scale from 1 (low) to 5 (high). Participants rate the match of personalized driving performance, system reliability, level of trust, and understandability of the system's user interface after each trip. Similar to the previous section, participants are unaware of which system they were using. The results are displayed in Fig. \ref{fig:median}.


After collecting data, we conduct a Wilcoxon signed-rank test \cite{wilcoxon1992individual}, a robust non-parametric approach for paired data, to investigate the effectiveness of the VLM-based method. The null hypothesis ($H_0$) posits that the median of the baseline is greater than or equal to that of our method, while the alternative hypothesis ($H_1$) suggests that the median of the baseline is less than that of our method. Using a significance level of $p<.05$ to reject $H_0$, our analysis shows that the VLM-based method significantly outperforms the baseline across multiple metrics, including matching personalized driving behaviors and demonstrating greater reliability. In more challenging scenarios (e.g., lane changes, left turns), our VLM-based method's advantages are particularly more evident, leading to increased trust and improved understandability compared to the baseline.

\subsection{Ablation Study on Memory Module}

\begin{figure}[t!]
    \centering
    \includegraphics[width=0.9\linewidth]{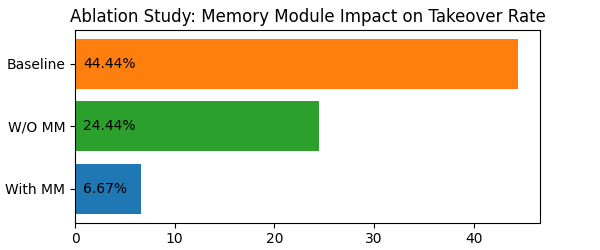}
    \caption{Effectiveness of memory modules in takeover rates}
    \vspace{-7mm}
    \label{fig:rag_takeover}
\end{figure}

To further validate the effectiveness of our RAG-based Memory Module (MM),  we conduct an ablation study with three human participants, comparing three configurations: our complete system with MM, the system without MM (W/O MM), and the baseline. The results demonstrate the significant impact of the MM on reducing takeover rates.

As shown in Fig. \ref{fig:rag_takeover}, With three participants, our complete system with the memory module achieves the lowest average takeover rate of 6.67\%. When we remove the memory module while keeping other components the same, the average takeover rate increases substantially to 24.44\%. The baseline system performs worst with a 44.44\% average takeover rate. These results indicate a 72.7\% reduction in takeover rate when adding the memory module to our base architecture and an 85\% overall reduction compared to the baseline. This significant improvement suggests that the RAG-based memory module plays a crucial role in maintaining personalized vehicle control by effectively utilizing historical interactions and user preferences.

\section{Conclusion}
\label{sec:conclu}

In this paper, we presented an on-board VLM-based framework designed to enhance motion control tasks in autonomous driving, offering a more human-centric and responsive user experience. Our personalized motion control system represents a novel integration of VLMs into autonomous driving, offering three significant contributions to human-centric driving experiences. First, through its RAG-based memory module, the system demonstrates advanced personalization capabilities, effectively learning and adapting to individual driving preferences while maintaining safety and comfort standards across various scenarios. Second, the fine-tuned VLM-based framework leverages multimodal reasoning to understand both complex visual scene information and implicit natural language instructions, enabling human-like human-vehicle interaction. Finally, the system achieves remarkable computational efficiency with an optimized 9B VLM implementation, processing with an average 1.6-second latency and less than 16 GB GPU memory consumption on standard vehicle hardware, making it feasible for commercial deployment. Through experiments, we demonstrated that our VLM-based approach enhanced driving reliability, trustworthiness, and personalization, reducing the need for human takeover by up to 76.8\% while maintaining near real-time response. This framework contributed to enhancing personalized autonomous driving by aligning vehicle behavior with individual user preferences and considering environmental information, marking a significant step toward a more adaptable, user-centered human-autonomy teaming solution.


{
    \small
    \bibliographystyle{ieeenat_fullname}
    \bibliography{main}
}


\end{document}